\documentclass{article}

\usepackage[nonatbib, final]{nips_2016} 

\usepackage[utf8]{inputenc} 
\usepackage[T1]{fontenc}    
\usepackage{hyperref}       
\usepackage{url}            
\usepackage{booktabs}       
\usepackage{amsfonts}       
\usepackage{nicefrac}       
\usepackage{graphicx}
\usepackage{subfig}
\usepackage{todonotes}
\usepackage{microtype}      
\usepackage{wrapfig}
\usepackage{comment}

\author{
  Amy Zhang\thanks{Authors are of equal contribution} \\
  Facebook \\
  \texttt{amyzhang@fb.com} \\
  \And
  Xianming Liu\footnotemark[1] \\ 
  Facebook \\
  \texttt{xmliu@fb.com} \\
  \And
  Andreas Gros \\
  Facebook \\
  \texttt{andreasg@fb.com} \\
   \And
  Tobias Tiecke \\
  Facebook \\
  \texttt{ttiecke@fb.com} \\
}

\begin{document}

\title{Building Detection from Satellite Images on a Global Scale}
\maketitle

\begin{abstract}
In the last several years, remote sensing technology has opened up the possibility of performing large scale building detection from satellite imagery.  Our work is some of the first to create population density maps from building detection on a large scale.  The scale of our work on population density estimation via high resolution satellite images raises many issues, that we will address in this paper.  The first was data acquisition.  Labeling buildings from satellite images is a hard problem, one where we found our labelers to only be about 85\% accurate at.  There is a tradeoff of quantity vs. quality of labels, so we designed two separate policies for labels meant for training sets and those meant for test sets, since our requirements of the two set types are quite different.  We also trained weakly supervised footprint detection models with the classification labels, and semi-supervised approaches with a small number of pixel-level labels, which are very expensive to procure.  
\end{abstract}

\section{Introduction}
With the recent improvements in remote sensing technology, there has been a lot of work in building detection and classification from high resolution satellite imagery.  However, we are the first to implement a system on a global scale.  Other work uses handpicked features to define buildings \cite{DBLP:journals/corr/Cohen0KCD16} \cite{autorooftop} which would not scale well across countries with very different styles of buildings. The work closest to ours is done by Yuan \cite{DBLP:journals/corr/Yuan16}, which also uses pixel level convolutional neural networks for building detection, but is only validated on a handful of cities in the US and likely would not transfer well to smaller settlements or other countries.    

In order to speed up our pipeline we need a fast bounding box proposal algorithm to limit the number of images that need to be run through our convolutional neural network.  To maintain high recall, however, we need to be careful to not filter out too many candidates.  We used a naive bounding box proposal algorithm, by performing straight edge detection to extract smaller masks to run through our classification network.  This reduced the amount of landmass to process by 50\%. The distribution of buildings is still very negatively skewed, where only 2\% of proposals are positive.  This also means we need to sample a large number of masks in order to have confident precision and recall numbers by country.  We also use a weak building classifier to filter masks with over 0.3 IoU (intersection over union) by choosing the mask with the highest probability of containing a building in the center, since these overlapping masks are likely to contain the same building.  

Discovering systematic issues with our models is also a slow, manual problem that requires visualization of .kmz files, pinpointing large numbers of false positive or false negative areas, and debugging the causes.   The problems encountered included noise, contrast issues, cloud cover, or just deficiencies in the model, and we set up a feedback loop to fix those problems. 

We will be open sourcing our population density results as well as our labeled dataset as a benchmark for future efforts.

\section{Dataset Collection Issues}
We have two goals for data collection, obtaining labels for training, and accuracy numbers on a country level.  Obtaining accuracy numbers of the entire pipeline for a single country requires randomly sampling from all possible 64x64 masks.  That distribution is incredibly skewed, and randomly sampling enough masks to obtain a reasonable confidence interval on accuracy is expensive.  Instead, we measure how well our neural network performs building classification by randomly sampling from the distribution of masks generated by our bounding box proposal algorithm.  The assumption is that the bounding box proposal algorithm only eliminates clear negatives, so reduces skew on the underlying distribution without affecting recall of the overall pipeline.  This drops the number of labels we need by a factor of 10, because our new distribution now is 2\%-5\% positive.

Collecting a training set went through several iterations because we want a more balanced dataset for training so the model can get enough samples of both the background and the building classes.  We also employ simple active learning techniques by sampling from masks the network was "less sure" about, where the probability was closer to the threshold.  

\section{Generalizing a Global Model}
Training a global building classification model has trade-offs.  Buildings can look very different across different countries, but there is still a lot of information that can be transferred from country to country.  We initially started with a model trained only on Tanzania, which when applied to a new country had a large drop in accuracy.  However, we found that as we labeled data in more countries and re-trained our model with the new data, our new global model performed better on Tanzania than a Tanzania specific model.  The generalizations learned from other countries made the model more robust.

Another argument for training a global model is that building a large training set takes time, and the amount of data required to train a model from scratch for each country was prohibitive.  

The trade-off is that the global model doesn't work equally well on all countries, and we found it necessary to perform some amount of model specialization.  We fine-tuned the global model with the same samples it had seen from the initial training, but only from a handful of countries that we wanted it to improve upon.  We saw gains of 20-40\% in precision and recall on the validation set using the extra fine-tuning step, but noticed there were trade-offs.  The training and validation sets gave no evidence of overfitting, but we saw an increase in systematic false positives when visualizing the results on a country level, in certain countries.  

\subsection{Building Classification Model}
The classification model we trained was a weakly supervised version of SegNet \cite{DBLP:journals/corr/BadrinarayananK15}, which is a fast yet accurate pixel classification network that uses deconvolution layers.  We trained with weak ``pixel level'' labels, and generate a mask level probability using global average pooling on the final pixel level probabilities over the 64x64 mask. We have 500TB of satellite imagery, and being able to run the model over all these countries (multiple times) is crucial for fast iteration.  It was a non-trivial task to develop a model that was large enough to capture the complex idea of what defines a building, while also being small enough to run quickly during inference time.  SegNet performed well on this by saving the indices from the max pooling layers to perform non-linear upsampling in the deconvolution layers. 

\subsection{Building Segmentation Model}

\begin{figure}[!h]
   \centering
   \includegraphics[width=0.5\textwidth]{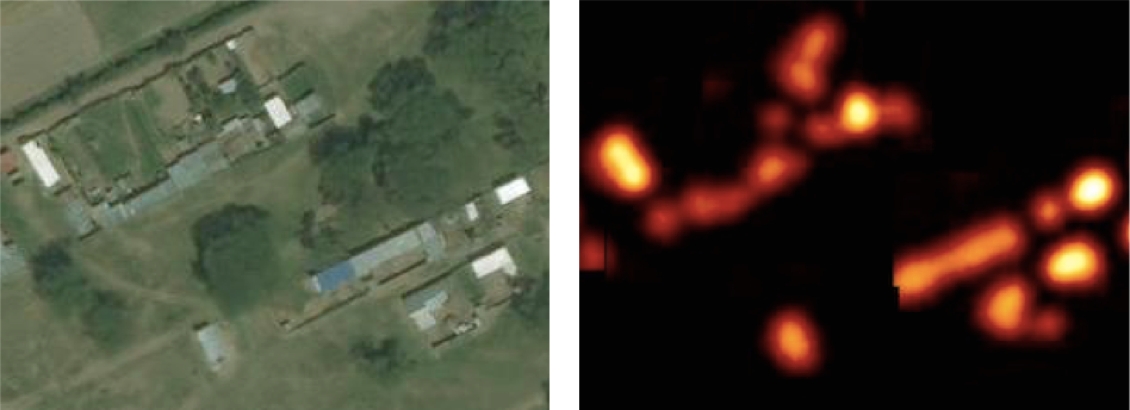}
   \caption{Semantic segmentation results using weakly-supervised model.}
   \label{fig:segmentation}
\end{figure}

Finely pixel-wise labeled data is extremely time consuming to acquire, and errors will accumulate especially for small foreground objects. Instead of utilizing fully supervised semantic segmentation method such as FCN \cite{long2015fully}, we investigated weakly supervised segmentation models relying on feedback neural network \cite{cao2015look}, which utilizes the large amount of ``cheap'' weakly-supervised training data. Notably, to increase the efficiency of semantic segmentation, the classification model is composed to filter out negative candidate regions. By combining results from both models, the segmentation model successfully suppress false positives and generate best results, with an example shown in Figure~\ref{fig:segmentation}

\section{Dealing with Systematic Errors}

\subsection{Finding Systematic Errors}

The precision and recall numbers we measure by randomly sampling from the mask candidates do not account for systematic errors arising from varying satellite image quality. To discover those systematic errors, we adopt both visually inspection and evaluation using external data.

Intuitively, we visualize our results by construction \emph{KMZ} files and overlaying with Google Earth to manually pinpoint areas of concern. We also use this strategy to sample \emph{ambiguous} training data to fine-tune our model to reduce the chance of further systematic errors.
Moreover, we also quantitatively measure systematic errors at a coarser scale by comparing our results with external datasets on those areas with adequate data coverage. However, it is still an open question to discover systematic errors on large scale with less manual work.

\subsection{Data Quality}

\begin{figure}[h!] 
    \centering
    \subfloat[Before Denoising]{%
        \includegraphics[height=5cm]{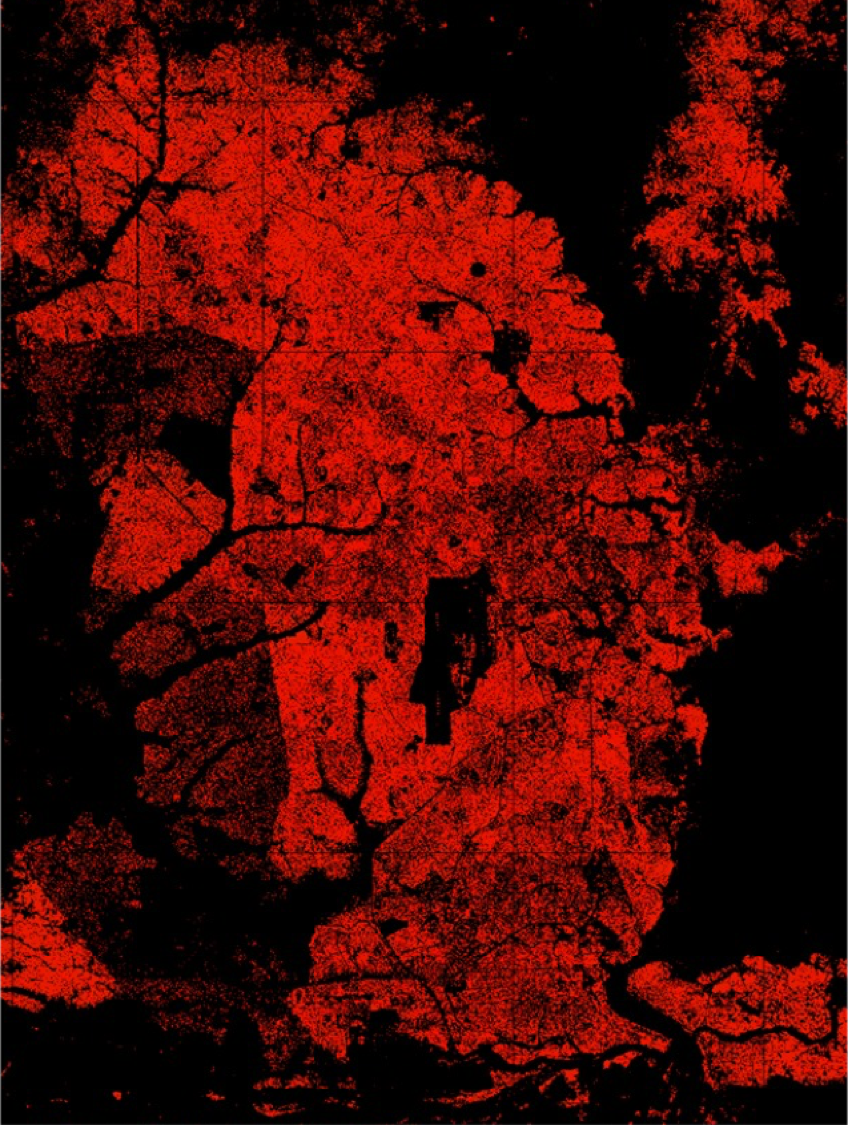}%
        }%
    \subfloat[After Denoising]{%
        \includegraphics[height=5cm]{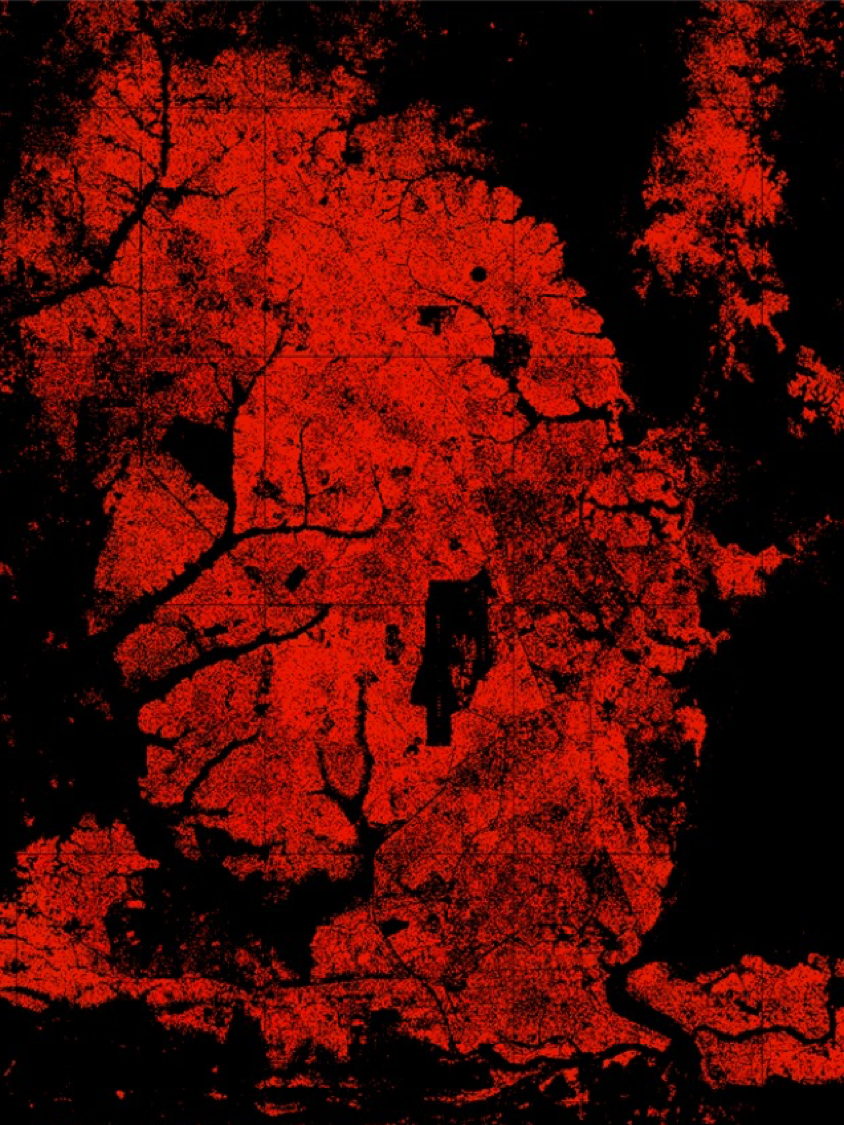}%
        }%
    \caption{Classification Results before and after denoising.}
    \label{fig:denoising}
\end{figure}

One of the reasons for systematic errors is also issues with data quality. The satellite images are taken at various times of day, and pre-processed across multiple layers for the highest quality image.  However, areas with a lot of cloud cover tend to have much fewer clear images taken, and so quality suffers.  This has an impact on our model, since most of the data is randomly or semi-randomly sampled, and so it does not get a lot of exposure to these poorer quality images during training. We use geographical meta-information to further detect the cloud occlusion during deploying stage.

Another key factor of low data quality comes from noise, which are introduced in either imaging or image enhancing phases. Traditional image denoising approach such as BM3D \cite{dabov2006image} is computationally expensive in handling large imagery files, and can only work for limited type of noises, such as white noise. To this end, we train a shallow neural network end-to-end by mimicking several kinds of noise existed in satellite images. The trained denoising model is appended as a transformer before imagery is fed to the classification network.
Comparison of classification results of the same low data quality area before and after denoising is shown in Figure \ref{fig:denoising}.

\section{Results}
Overall the SegNet model by itself achieves a precision and recall of $pr=0.9$, $re=0.89$ on a global dataset where the imbalance is such that $93\%$ of the randomly sampled testing data is not a building.
Below we have some heat maps generated of building density in three countries: Mozambique, Madagascar, and India. 

\begin{figure}[h!]
    \centering
    \subfloat[Mozambique]{%
        \includegraphics[height=5cm]{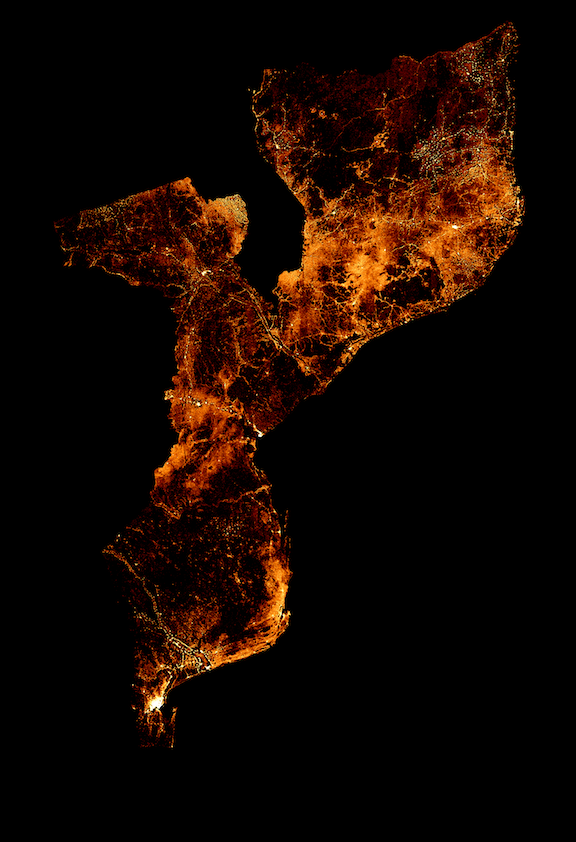}%
        \label{fig:a}%
        }%
    \subfloat[Madagascar]{%
        \includegraphics[height=5cm]{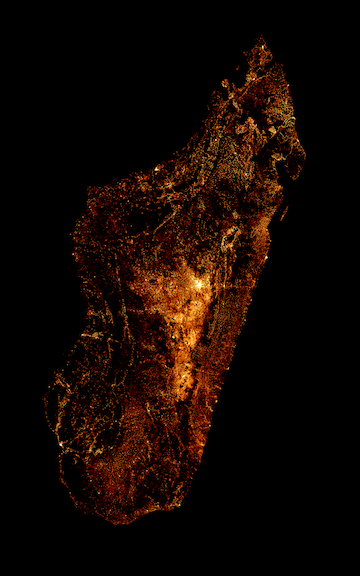}%
        \label{fig:b}%
        }%
    \subfloat[India]{%
        \includegraphics[height=5cm]{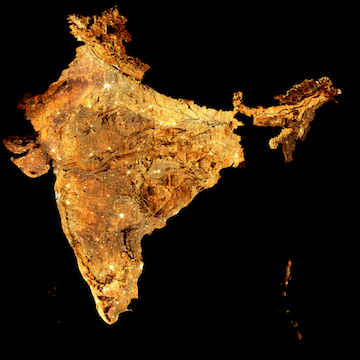}%
        \label{fig:c}%
        }%
    \caption{Building Heat Maps}
\end{figure}

So far we have released datasets for 5 countries: Haiti, Malawi, Ghana, South Africa, and Sri Lanka.  The rest are pending validation with third party groups.  Below we show precision recall curves and best F-score with confidence intervals for each of the countries released.   
\begin{figure}[h!]
    \centering
    \subfloat[Pr/Re Curves]{%
        \includegraphics[height=5cm]{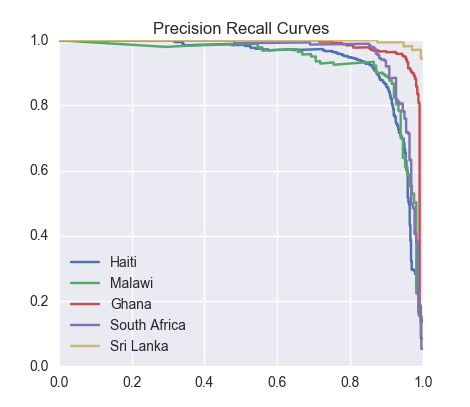}%
        \label{fig:a}%
        }%
    \subfloat[Confidence Intervals for F-Score]{%
        \includegraphics[height=5cm]{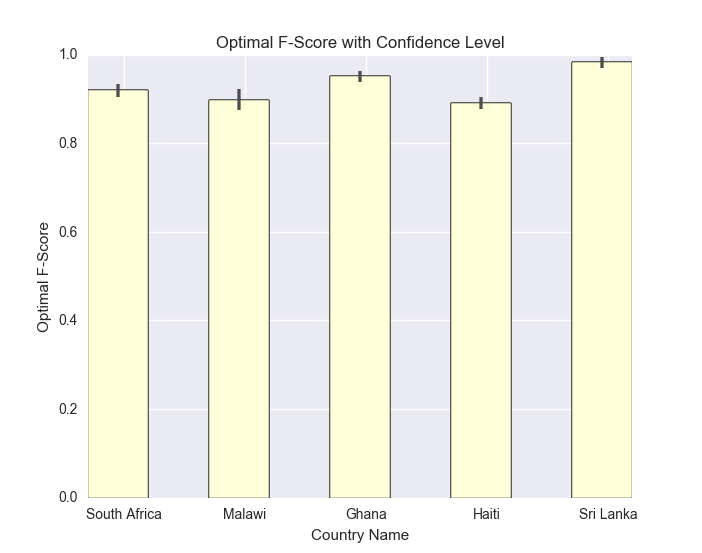}%
        \label{fig:b}%
        }%
    \caption{Classification Performance}
\end{figure}

\begin{figure}[h!] 
   \centering
   \includegraphics[width=0.5\textwidth]{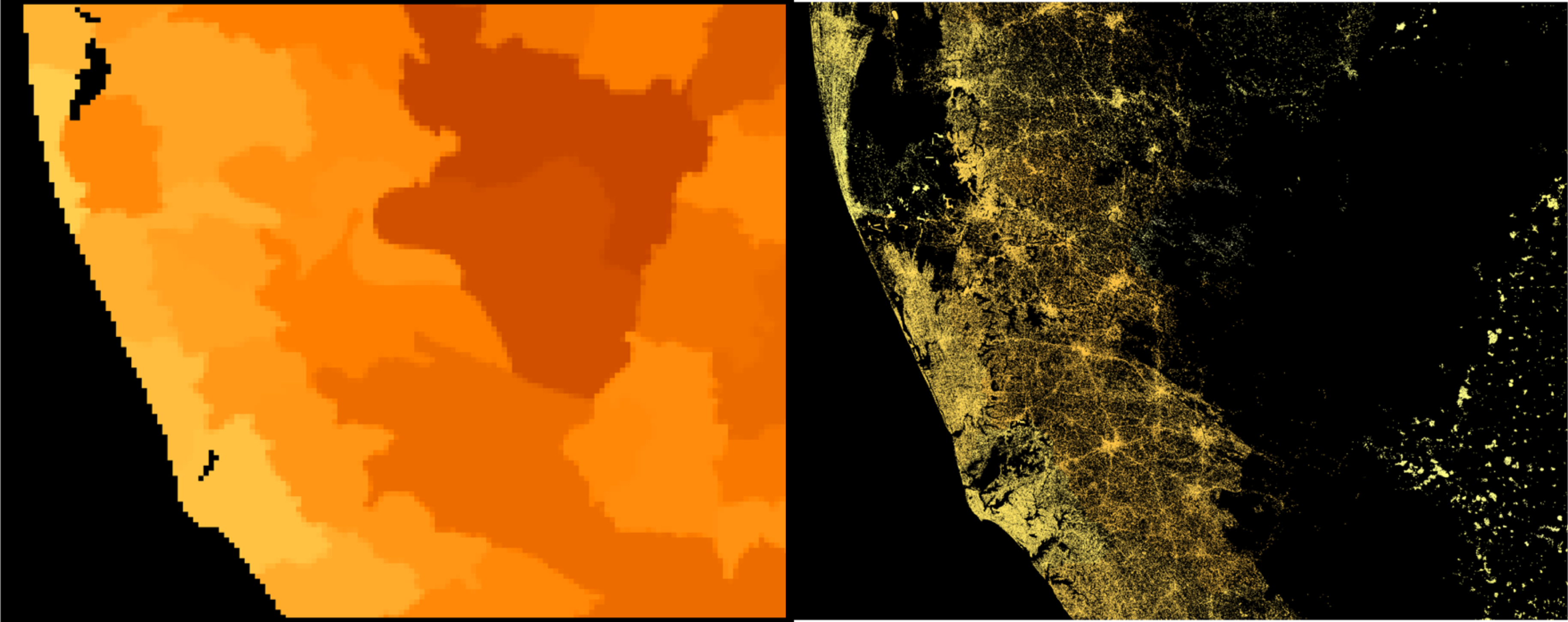}
   \caption{Comparison of Galantis and our results}
   \label{fig:stateofart}
\end{figure}

The estimation of population density via settlement buildings as a proxy results in significant improvement compared with previous efforts. Figure~\ref{fig:stateofart} shows the comparison of previous highest resolution estimation from Galantis and our own results. This gives a totally new perspective to various social / economic research.

\section{Conclusion}
We have built one of the first building detection systems that can be deployed at a global scale.  Future work includes reducing the amount of iteration required to achieve a robust model as we roll out to more countries, the biggest problem of which is detecting systematic errors.  Detecting and solving these systematic issues in classification is still a work in progress.  We are still looking into ways to automate the data validation process and data collection methods further, which will also shorten the length of each iteration required to improve our dataset accuracy.  

\small
\bibliographystyle{unsrt}
\bibliography{egbib}

\begin{thebibliography}{1}

\bibitem{DBLP:journals/corr/Cohen0KCD16}
Joseph~Paul Cohen, Wei Ding, Caitlin Kuhlman, Aijun Chen, and Liping Di.
\newblock Rapid building detection using machine learning.
\newblock {\em CoRR}, abs/1603.04392, 2016.

\bibitem{autorooftop}
M.~Cote and P.~Saeedi.
\newblock Automatic rooftop extraction in nadir aerial imagery of suburban
  regions using corners and variational level set evolution.
\newblock {\em IEEE Transactions on Geoscience and Remote Sensing},
  51(1):313--328, 2013.

\bibitem{DBLP:journals/corr/Yuan16}
Jiangye Yuan.
\newblock Automatic building extraction in aerial scenes using convolutional
  networks.
\newblock {\em CoRR}, abs/1602.06564, 2016.

\bibitem{DBLP:journals/corr/BadrinarayananK15}
Vijay Badrinarayanan, Alex Kendall, and Roberto Cipolla.
\newblock Segnet: {A} deep convolutional encoder-decoder architecture for image
  segmentation.
\newblock {\em CoRR}, abs/1511.00561, 2015.

\bibitem{long2015fully}
Jonathan Long, Evan Shelhamer, and Trevor Darrell.
\newblock Fully convolutional networks for semantic segmentation.
\newblock In {\em Proceedings of the IEEE Conference on Computer Vision and
  Pattern Recognition}, pages 3431--3440, 2015.

\bibitem{cao2015look}
Chunshui Cao, Xianming Liu, Yi~Yang, Yinan Yu, Jiang Wang, Zilei Wang, Yongzhen
  Huang, Liang Wang, Chang Huang, Wei Xu, et~al.
\newblock Look and think twice: Capturing top-down visual attention with
  feedback convolutional neural networks.
\newblock In {\em Proceedings of the IEEE International Conference on Computer
  Vision}, pages 2956--2964, 2015.

\bibitem{dabov2006image}
Kostadin Dabov, Alessandro Foi, Vladimir Katkovnik, and Karen Egiazarian.
\newblock Image denoising with block-matching and 3d filtering.
\newblock In {\em Electronic Imaging 2006}, pages 606414--606414. International
  Society for Optics and Photonics, 2006.

\end{thebibliography}

\end{document}